\documentclass[a4paper,fleqn]{cas-dc}

\usepackage[numbers]{natbib}
\usepackage{subfigure}

\def\tsc#1{\csdef{#1}{\textsc{\lowercase{#1}}\xspace}}
\tsc{WGM}
\tsc{QE}
\tsc{EP}
\tsc{PMS}
\tsc{BEC}
\tsc{DE}

\begin{document}
\let\WriteBookmarks\relax
\def\floatpagepagefraction{1}
\def\textpagefraction{.001}
\shorttitle{Computers \& Security}  
\shortauthors{H. Wang et~al.}

\title [mode = title]{An Efficient Pre-processing Method to Eliminate Adversarial Effects}                      


\author{Hua Wang}

\credit{Conceptualization of this study, Methodology, Software}

\address{Key Lab of Intelligent Computing and Signal Processing of Ministry of Education, School of Computer Science and Technology, Anhui University, Hefei 230601, China}

\author{Jie Wang}

\author{Zhaoxia Yin}[orcid=0000-0003-0387-4806]
\cormark[1]



\cortext[cor1]{Corresponding author.\\
E-mail address: yinzhaoxia@ahu.edu.cn (Z. Yin).}

\cortext[cor2]{This research work is partly supported by National Natural Science Foundation of China (61872003, U1636206).}

\begin{abstract}
Deep Neural Networks (DNNs) are vulnerable to adversarial examples generated by imposing subtle perturbations to inputs that lead a model to predict incorrect outputs. Currently, a large number of researches on defending adversarial examples pay little attention to the real-world applications, either with high computational complexity or poor defensive effects. Motivated by this observation, we develop an efficient preprocessing method to defend adversarial images. Specifically, before an adversarial example is fed into the model, we perform two image transformations: WebP compression, which is utilized to remove the small adversarial noises. Flip operation, which flips the image once along one side of the image to destroy the specific structure of adversarial perturbations. Finally, a de-perturbed sample is obtained and can be correctly classified by DNNs. Experimental results on ImageNet show that our method outperforms the state-of-the-art defense methods. It can effectively defend adversarial attacks while ensure only very small accuracy drop on normal images.

\end{abstract}


\maketitle

\section{Introduction}
\begin{figure} 
\centering
\includegraphics[width=3.6in,height=1.30in]{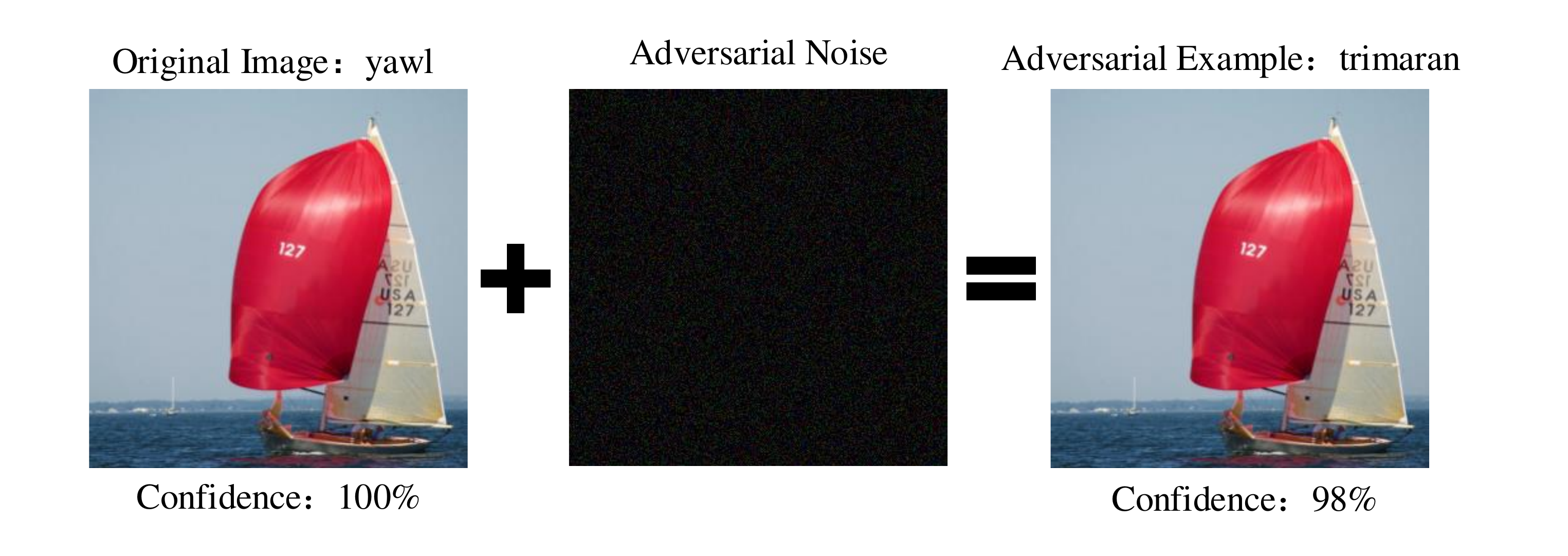} 
\caption{The generation process of adversarial example.} \label{img1}
\end{figure}

Deep learning \cite{Xing2016Deep} \cite{nguyen2018auto} has made great breakthroughs in the field of computer vision \cite{ni2018malware}, e.g., improving the image recognition accuracy to the human level. The emerging technology solves the problems inherent in traditional machine learning \cite{pedregosa2011scikit} and artificial intelligence \cite{russell2016artificial}. However, Szegedy et al. \cite{Szegedy2013Intriguing} first discovered an intriguing weakness of deep neural networks in image classification. As reported, the neural network can predict a wrong classification result by deliberately applying imperceptible adversarial noises to an image, thus raising the concept of the adversarial example. As shown in Fig. \ref {img1}, left picture can be normally classified by DNNs, whereas right picture is misclassified after adding a certain perturbations. Although they are visually similar to each other, DNNs might still output the wrong prediction with high confidence. The existence of adversarial examples not only poses a huge threat to the application of DNNs in security sensitivity \cite{navarro2018systematic}, but also bring certain security risks to the researches, such as automatic driving \cite{zhong2017class} and identity recognition \cite{irons2017face} \cite{wang2017bimodal}.

To better comprehend DNNs and overcome the above drawbacks, the researchers have proposed a series of schemes to defend against various existing attacks such that enhancing the robustness of DNNs. However, current researches on defending adversarial examples pay little attention to the actual situation and often ignore the high computation cost. Moreover, the researches on defensive methods have lagged behind the ones on attack methods. It is necessary to develop more efficient and practical approaches to defend against the attack of adversarial examples.

\begin{figure*} 
\centering
\includegraphics[width=7in]{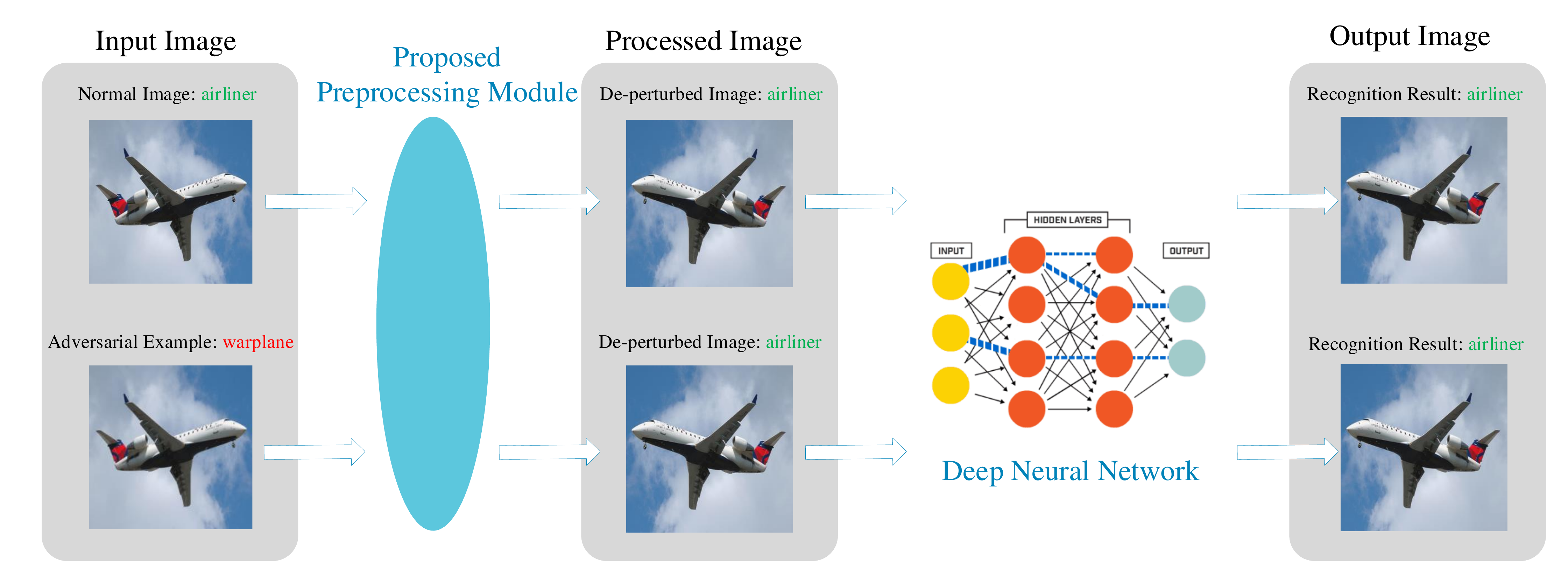} 
\caption{The overall framework of the proposed method.} \label{img2}
\end{figure*}

Based on above reasons, we develop a defense method based on two simple image transformations. Due to the weak generalization of iterative attacks, the specific structure of adversarial perturbations can be destroyed by low-level image transformations, e.g., compression, flipping. To be specific, we combine WebP compression \cite{lian2012webp} and flipping to weaken adversarial attacks. The general framework of the proposed scheme is shown in Fig. \ref{img2}. First, both normal image and adversarial image are preprocessed by our method and then the two de-perturbed images obtained by processing are fed into the neural network to classify. In the end, both the adversarial image and the normal image can be correctly classified by the model.

As we all know, JPEG compression \cite{Raid2014Jpeg} \cite{ozah2019compression} is an effective way to defend adversarial examples. However, in the case of medium and low bitrates, JPEG compression is prone to blocking artifacts \cite{singh2007reduction}, resulting in poor image quality and loss of classification accuracy. WebP lossy compression is specially designed to reduce the image details that are difficult to be perceived by human beings to compress the image volume, which can effectively remove imperceptible perturbations in adversarial examples. Forthemore, WebP compression introduces loop filtering \cite{ginesu2012objective}, which can eliminate the block effect. It not only can effectively destroy adversarial structure, but also ensure that the compressed image quality still high.

Numerous experiments are carried out on the ImageNet \cite{krizhevsky2012imagenet}. Experimental results show that WebP compression is superior to JPEG compression in terms of defending adversarial attacks. WebP compression and flipping are first applied to adversarial defense, combining the two image transformations can provide more excellent defensive effects. It can even defend the most advanced white-box iterative attack methods. Compare with the state-of-the-art defense methods: Comdefend \cite{jia2019comdefend} and JPEG compression \cite{das2017keeping}, the top1 accuracy of model of our method is more than 20\% higher than Comdefend and 5\% higher than JPEG compression.

The rest of this paper is organized as follows: In Sect. \ref{tit2}, we discuss related works on attack methods and defensive methods. In Sect. \ref{tit3}, the proposed scheme is described in detail. Sect. \ref{tit4} shows the experimental setup as well as experimental results and analysis. The conclusion is given in Sect. \ref{tit6}.

\section{Related works \label{tit2}}
In this section, we will review related works from two aspects: the attack methods of generating adversarial examples, and the defensive methods of resisting adversarial examples.

\subsection{Attack methods}
\subsubsection{Fast Gradient Sign Method(FGSM) \cite{Goodfellow2014Explaining}}
Goodfellow et al. proposed the Fast Gradient Sign Method (FGSM), a way of rapidly generating adversarial examples. Given the input image, maximum direction of gradient change of the deep learning model is found, and adversarial disturbances are added in this direction, resulting in the wrong classification result. The FGSM adds disturbances to the image by increasing the image classifier loss. The generated adversarial example is formulated as follows:
\begin{equation}
x^{adv}=x + \epsilon \cdot sign(\bigtriangledown_x J(x,y))
\end{equation}
where $J(x,y)$ denotes the cross entropy cost function, $x$ is the input image, $y$ is the true label of the input image, and $\epsilon$ is the hyperparameter that determines the magnitude of the disturbances.

\subsubsection{IFGSM \cite{kurakin2016adversarial}}
IFGSM was proposed as an improved version of FGSM, in which the perturbations affected by $L\infty$ constraints could be calculated. This method applied FGSM multiple times with small disturbances instead of applying a large disturbance noise. The pixels are appropriately clipped after each iteration to ensure that the results remain in the neighborhood of the input image $x$.
\begin{equation}
x^{(i)}=clip_{x,\epsilon}(x^{(i-1)} + \epsilon \cdot sign(\bigtriangledown_{x^{(i-1)}} J(x^{(i-1)},y)))
\end{equation}

\subsubsection{DeepFool \cite{moosavi2016deepfool}}
Moosavi-Dezfooli et al. proposed the DeepFool attack algorithm, which was used to calculate the minimum adversarial disturbances. The DeepFool was a untargeted attack method that generated an adversarial example by iteratively perturbing the image. First, it explored the nearest decision boundary, and then the image was slightly modified to reach this boundary in each iteration. The algorithm might not stop running until the modified image changed the classification result. Compared with FGSM, the disturbances generated by this method are smaller when the fooling rate is similar. The resulting disturbances are more difficult to detect.

\subsubsection{Carlini and Wagner(C\&W) \cite{carlini2017towards}}
The C\&W algorithm was proposed by Carlini and Wagner. The attack can be a targeted attack or a untargeted attack, and the distortion was measured by three measures: $(L_0, L_2, L\infty)$. It is more efficient than all previously known methods in terms of attack success rate achieved with minimal perturbation amounts. The untargeted $L_2$ norm attack version has the best performance. In this paper, we use this method to generate adversarial examples. CW\_L2 is an optimization - based attack method that generates adversarial examples by solving the following optimization problems:

\begin{equation}
  min\Vert x-x^{'}\Vert^2 +  \lambda max(-\kappa, Z(x^{'})_\kappa - max\{ Z(x^{'})_{\kappa^{'}} : \kappa^{'}\neq \kappa\})
\end{equation}
Where $\kappa$ controls the confidence that the image is misunderstood by the model, i.e., the confidence gap between the sample category and the real category, we set $\kappa$ = 0 in this paper. $Z(x^{'})_{\kappa^{'}}$ is the logical output of the category $\kappa^{'}$.

\begin{figure*}  
\centering
\includegraphics[height=2.0in]{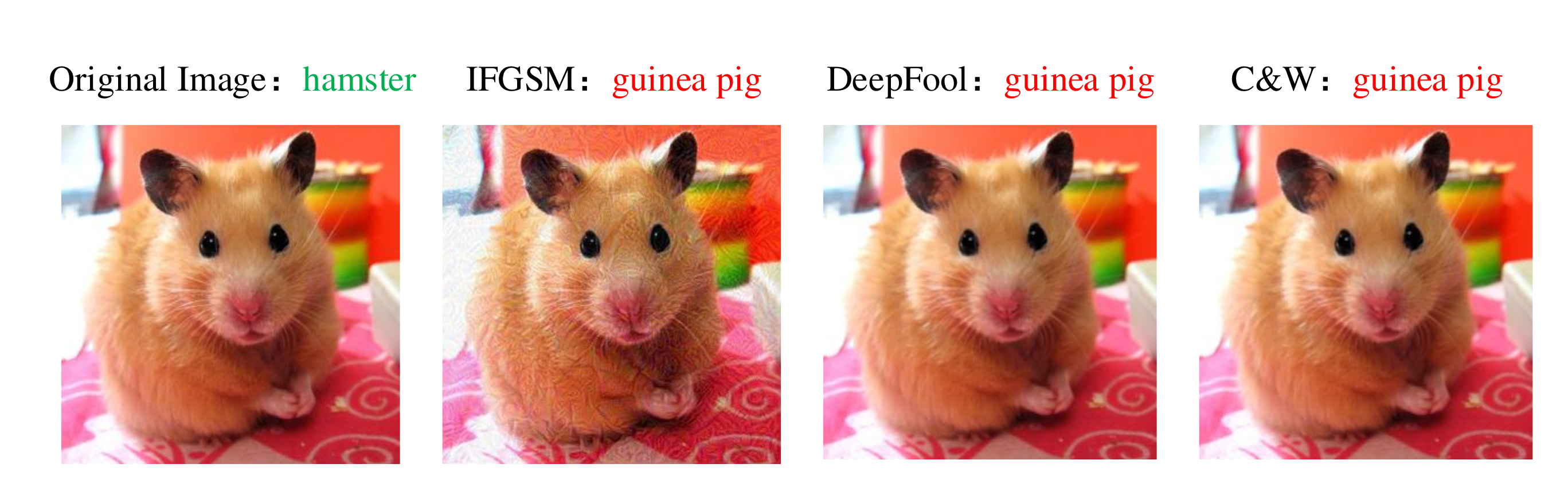} 
\caption{The adversarial examples generated on Inception\_v3. The image on the left is the original image, the other three images are adversarial examples generated by IFGSM ,DeepFool and C\&W respectively.} \label{img8}
\end{figure*}

As shown in Fig. \ref{img8}, adversarial images generated using IFGSM ,DeepFool and C\&W. We can see that the adversarial perturbations generated by the IFGSM attack algorithm are relatively large, and the human eye can perceive subtle disturbances. The adversarial examples generated by DeepFool and C\&W are too small to make a difference from the original image. Moreover, these two attack algorithms are the strongest attacks with a high attack success rate.

\subsection{Defensive methods}
Currently, defensive measures of adversarial examples are mainly divided into the following two categories.

\subsubsection{Modifying the neural network}
Adversarial Training \cite{tramer2017ensemble} was proposed as a typical method to defend against adversarial examples. It used adversarial examples as part of the model training set and worked with the original samples to train the model. Gradient Masking \cite{papernot2017practical} was also proposed to modify the model gradient to enhance the robustness of the model. By hiding the gradient information on the model training, it is difficult for the attack algorithm to attack the model through the gradient solution method. However, these two methods require a large amount of training data, and the training process is complicated and time consuming.

\subsubsection{Modifying input data}
Song et al. claimed that the PixelDefend method \cite{song2017pixeldefend} could transform the perturbed image into a clean image before inputting the sample into the classifier. PixelDefend mainly removes the perturbations by simulating the spatial distribution of the image. When the space is too large, the simulation result is very poor. Xie et al. added two random transformation layers to the model \cite{xie2017mitigating}, in which one was for arbitrarily resizing the image and the other was for arbitrarily padding the image to reduce the attack effectiveness of the adversarial example. Liao et al. treated imperceptible perturbations as noises, and designed a high level representation guided denoiser (HGD) \cite{liao2018defense} to remove these noises. Obviously, HGD is effective because it does not require neural network retraining. However, when training denoiser, a large number of adversarial examples are required, i.e., it is difficult to obtain good HGD with only a small number of adversarial examples. Thang et al. reduced the fooling rates of the networks by rotating the adversarial images \cite{thang2019image}. Jia et al. proposed the Comdefend method \cite{jia2019comdefend}, which mainly characterized by compressing and reconstructing the images through the convolutional neural networks to destroy the disturbance structure of the adversarial pictures. Das et al. applied JPEG compression \cite{das2017keeping} to remove the perturbations in the adversarial examples.

\section{Approach \label{tit3}}
The goal of the defense is to make the neural network more robust to adversarial examples, i.e., it can classify adversarial images correctly with little performance loss on non-adversarial images. To achieve this goal, we propose a method based on two image transformations. 

\subsection{The implementation of proposed method}
\begin{figure*} 
\centering
\includegraphics[width=7.0in]{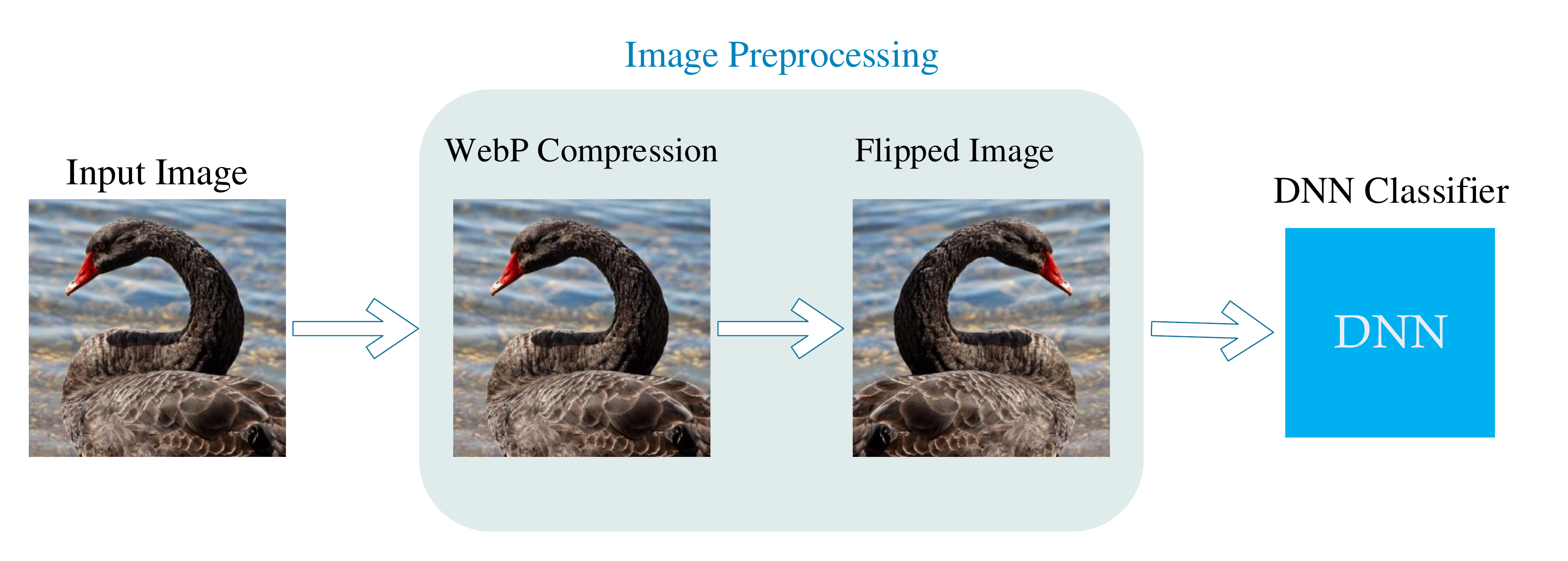} 
\caption{The processing of the proposed method.For an input image , before fed into the model , first use WebP compression , and then flip the compressed image .} \label{img3}
\end{figure*}

Due to the weak generalization of iterative attacks, low-level image transformations, e.g., compression, flipping, may probably destroy the specific structure of adversarial perturbations, thus, to make the best of both worlds, we combine WebP compression and flipping together to defend adversarial examples. The processing of the proposed method is shown in Fig. \ref{img3}. The order of WebP compression and flipping has no influence on the effect of defense.

WebP is Google's latest open source image format that supports lossy compression and lossless compression, derived from the image encoding format VP8. WebP lossy compression is specially designed to reduce the image details that are difficult to be perceived by human beings to compress the image volume, which can remove the small noises in adversarial images while ensuring that the compressed image quality is still high. In addition, conversions on both JPEG and PNG are excellent, stable, and consistent. We know that the most typical compression method, JPEG compression, which has an excellent effect on  defending adversarial attacks. However, it can produce blocking artifacts at low and medium bitrates, resulting in poor image quality. Since WebP compression introduces loop filtering, the block effect can be eliminated. Experiments show that the overall defense effect of WebP compression is better than JPEG compression. The compression degree is represented by the compression quality factor (QF), which ranges from 0 to 100. The greater degree of compression, the smaller quality factor, and the greater the difference between compressed image and its original image.

Flip operation only changes the position of pixels, but does not change the value of pixels. There are two ways to flip the image: Image. FLIP \_LEFT \_RIGHT, Image. FLIP \_TOP \_BOTTOM. In the experiment, we found that images are flipped by Image. FLIP \_LEFT \_RIGHT \footnote{https://www.programcreek.com/python/example/89936/pil.image. FLIP \_LEFT \_RIGHT}, and the classification accuracy of the model is higher, which is related to the data input method during model training, so we choose to flip the image by Image. FLIP \_LEFT \_RIGHT to defend adversarial examples. Goodfellow found that even a small perturbation in a linear high-dimensional space can have a big impact on the output \cite{Goodfellow2014Explaining}. The original image $x$, perturbed $\eta$, adversarial example: $x^{adv}=x + \eta$, but now consider adding the weight vector $\omega$, then there is a formula: $\omega^{T}x^{adv}=\omega^{T}x + \omega^{T}\eta$. The adversarial disturbances affect the activation function by $\omega^{T}\eta$. Assume that the width of the original image is $width$, and the pixels of $(m, n)$ points become $(width-m, n)$ after a left-right flip, $\omega^{T}\eta$ will be changed so that invalidates adversarial attacks.

\section{Experiments \label{tit4}}
\subsection{Experimental Setup}
\begin{itemize}
\item {\bfseries Model:} Mainly do experiments on trained inception\_V3 and ResNet101 \footnote{https://pytorch.org/docs/stable/torchvision/models.html}.
\item {\bfseries Dataset:} In the experiments, we selected images from the ILSVRC 2012 verification set \footnote{http://www.image-net.org/challenges/LSVRC/2012/nonpub-downloads}, which consists of 50,000 images and contains 1000 classes. Since it is meaningless to attack misclassified images, we randomly selected 5000 images that can be correctly classified by inception\_V3 and ResNet101, and called them as benign images in this paper.
\item {\bfseries Attack method:} We choose the most advanced three attack algorithms to generate adversarial examples: IFGSM, DeepFool, C\&W. 
\item {\bfseries Evaluation:} We use the model's top1 accuracy to evaluate the performance of defense methods.
\end{itemize}

\subsection{Experimental Results and Analysis }

\subsubsection{JPEG compression and WebP compression}
\begin{figure*} 
\centering
\includegraphics[width=7in]{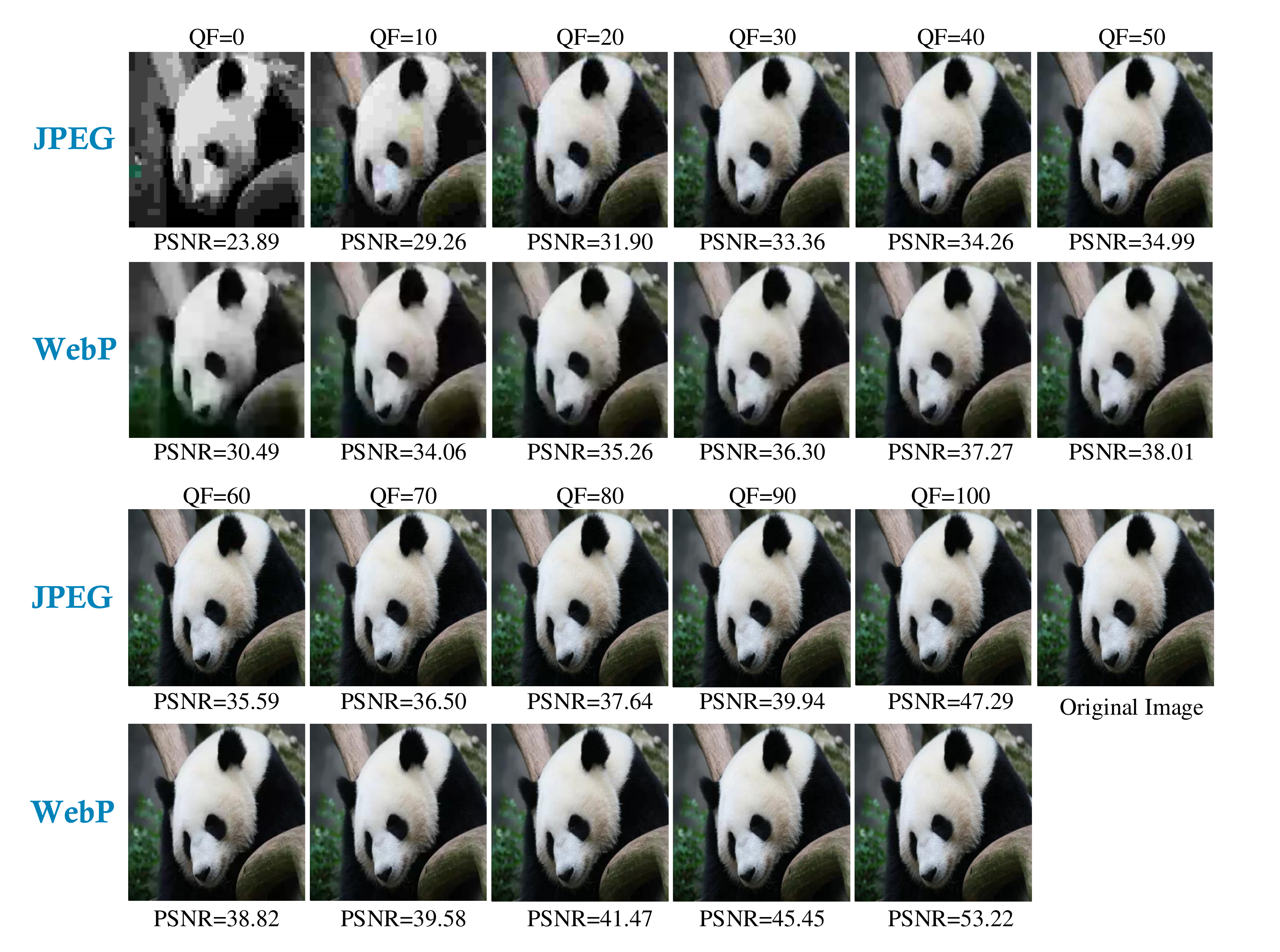} 
\caption{Comparison of the sample quality between JPEG and WebP at different quality factors. PSNR of the compressed image under each image.} \label{img4}
\end{figure*}

To compare the defense effects of JPEG compression and WebP compression. We take the most advanced attack methods to generate adversarial images, select successful adversarial examples(SAEs), and use WebP compression and JPEG compression for defense respectively. As shown in Table \ref{tab1}. When the quality factor ranges from [50, 100], the top1 accuracy of WebP compression is equivalent to JPEG compression. However, when the quality factor ranges between [0, 50], the top1 accuracy of WebP compression is 5.18\% higher than that of JPEG compression, so the defense effect of WebP compression is significantly better than that JPEG compression. This is because JPEG compression is prone to block effects at low and medium bit rates, result in loss of classification accuracy. But WebP compression introduces loop filtering to eliminate block effects and keep the compressed image quality high. The image quality obtained by the two compression methods is shown in Fig. \ref{img4}. The experimental results show that the image quality of WebP compression is better than JPEG compression, and it is more effective in defending adversarial attacks.

\begin{table}
\caption{Top 1 accuracy of two defense methods: JPEG compression and WebP compression on Inception\_v3.} \label{tab1}
\label{sphericcase}
\begin{tabular}{@{\extracolsep{\fill}}lrrrl@{}}
\toprule
Defense & \multicolumn{1}{c}{QF$\subset$[0,50]} & \multicolumn{1}{c}{QF$\subset$[50,100]}  \\
\midrule
JPEG & 71.21\% & 58.91\% \\
WebP  & 76.39\% & 58.58\% \\
\bottomrule
\end{tabular}
\end{table}

\subsubsection{Experimental results}
Then, to compare our method with the two methods used in our method respective effects. On Inception\_v3, we generate four sets of successful adversarial samples that can successfully attack model. Top 1 accuracy of each defense on adversarial examples and benign images are shown in Table \ref{tab2}. It also shows the parameters when WebP and Flip combine best, and we test top 1 accuracy of WebP compression with the same quality factor. From the second column, we can see that the top1 accuracy of the benign samples is still high, indicating that WebP compression and Flip will only cause a small accuracy drop for benign images. In face of adversarial examples, only a slight compression and flipping images can achieve a good defense effect. When the IFGSM parameters are set at $\epsilon$ = 5/225 and 8/225, the top1 accuracy of the model is as high as 85.34\% and 78.83\%, respectively. For DeepFool and C\&W\_L2, the defense effect is better, and top1 accuracy can reach 86.90\% and 89.80\%, respectively. The experimental results show that the combination of WebP compression and Flip not only significantly improves the defense effect, but also bring smaller image content distortion.

\begin{table*}
\caption{Top 1 accuracy of our proposed scheme and its two image transformations: WebP compression and flip operation on Inception\_v3.} \label{tab2}
\label{sphericcase}
\begin{tabular*}{\textwidth}{@{\extracolsep{\fill}}lrrrrrrl@{}}
\toprule
Defense & \multicolumn{1}{c}{Benign Images} & \multicolumn{1}{c}{IFGSM($\epsilon$=5/225)} & \multicolumn{1}{c}{IFGSM($\epsilon$=8/225)} & \multicolumn{1}{c}{DeepFool}  & \multicolumn{1}{c}{C\&W\_L2} \\
\midrule
WebP & 93.28\%(QF$\subset$[60,80]) & 48.50\%(QF=80) & 48.18\%(QF=60) & 74.48\%(QF=70)& 81.63\%(QF=70) \\
Flip & 94.67\% & 76.69\% & 56.20\% & 71.03\% & 79.59\% \\
WebP+Flip & 92.48\%(QF$\subset$[60,80]) & 85.34\%(QF=80) & 78.83\%(QF=60) & 86.90\%(QF=70) & 89.80\%(QF=70) \\
\bottomrule
\end{tabular*}
\end{table*}

\subsubsection{Performance Evaluation}
To quantitatively evaluate the performance of our defense scheme, we compare our method with Comdefend and JPEG compression on ResNet101. Top 1 accuracy of each defense on adversarial examples and benign images are shown in Table \ref{tab3}, and we display the corresponding parameters when the classification accuracy is the highest. From the second column, we can see that the benign samples can achieve the best classification accuracy on our method. The top1 accuracy of the adversarial examples generated by the three attack methods is 9.68\%, 2.58\%, and 0\%, respectively. In order to ensure that the data used by the each group are consistent, we have not selected all successful adversarial examples. The top1 classification accuracy of our method is more than 20\% higher than Comdefend and more than 5\% higher than JPEG compression. Experimental results show that our proposed scheme is superior to the most advanced defense methods.

\begin{table*}
\caption{Top 1 accuracy of each adversarial defense on ResNet101. } \label{tab3}
\label{sphericcase}
\begin{tabular*}{\textwidth}{@{\extracolsep{\fill}}lrrrrrl@{}}
\toprule
Defense & \multicolumn{1}{c}{Benign Images} & \multicolumn{1}{c}{IFGSM($\epsilon$=3/225,5/225)} & \multicolumn{1}{c}{DeepFool} & \multicolumn{1}{c}{C\&W\_L2}  \\
\midrule
No Defense & 100\% & 9.68\% & 2.58\% & 0\% \\
Comdefend \cite{jia2019comdefend} & 79.78\% & 52.89\% & 64.77\% & 66.49\% \\
JPEG \cite{das2017keeping} & 93.14\%(QF$\subset$[40,60]) & 71.87\%(QF=50) & 79.29\%(QF=40) & 80.10\%(QF=60)  \\
Our Proposed & 95.84\% (QF$\subset$[80,90])& 80.12\%(QF=80) & 85.23\%(QF=90) & 86.59\%(QF=80) \\
\bottomrule
\end{tabular*}
\end{table*}

\section{Conclusion \label{tit6}}
To find an efficient method to resist the attack of adversarial examples, we combine two image transformations to mitigate adversarial effects. This method does not need to change the structure of the model. Before the samples are imported into the model, we perform WebP lossy compression and flip operation on the input images to destroy the specific structure of adversarial perturbations. The processed adversarial pictures are correctly recognized by the model like normal samples, achieving the purpose of defending against adversarial attacks.

The experimental results show that WebP compression method used in the proposed scheme has better defense effect than JPEG compression. And image flipping does not lose image features, but it can destroy the specific structure of adversarial perturbations to achieve the purpose of defense. The experimental results show that the effectiveness of combining these two methods seems surprising. It provides a high performance on defending against adversarial attacks while ensuring that the classification accuracy on the normal samples just decreases slight.

\bio{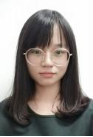}
{\bfseries Hua Wang} received her B.S. degree in software engineering from the school of computer science and technology, Anhui University of Technology, China, in 2018. She is currently pursuing the M.A. degree at the School of Computer Science and Technology, Anhui University. Her research interests include computer vision and AI security.
\endbio

~\\
~\\
~\\
~\\

\bio{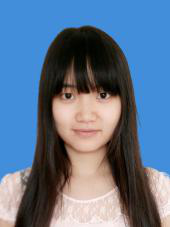}
{\bfseries Jie Wang} received the M.A. degree in the School of Computer and Information, Anqing Normal University in 2019. She is currently pursuing the Ph.D. degree at the School of Computer Science and Technology, Anhui University. Her main research interests include pattern recognition and AI security.
\endbio

~\\
~\\
~\\

\bio{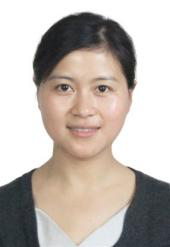}
{\bfseries Zhaoxia Yin}  received her B.Sc., M.E. \& Ph.D. from Anhui University in 2005, 2010 and 2014 respectively. She is a senior member of CSIG and member of CSIG Digital Media Forensics and Security Professional Committee. She is also an IEEE/ACM/CCF member and served CCF YOCSEF Hefei as an Associate Chair of the academic committee from 2016-2017. Currently she works as an Associate Professor and a Doctoral Tutor in School of Computer Science and Technology at Anhui University. She is also the Principal Investigator of two NSFC Projects. Her primary research focuses including Data Hiding, Privacy \& Security of  Multimedia  \& Machine Learning.
\endbio


\begin{thebibliography}{28}
\expandafter\ifx\csname natexlab\endcsname\relax\def\natexlab#1{#1}\fi
\providecommand{\url}[1]{\texttt{#1}}
\providecommand{\href}[2]{#2}
\providecommand{\path}[1]{#1}
\providecommand{\DOIprefix}{doi:}
\providecommand{\ArXivprefix}{arXiv:}
\providecommand{\URLprefix}{URL: }
\providecommand{\Pubmedprefix}{pmid:}
\providecommand{\doi}[1]{\href{http://dx.doi.org/#1}{\path{#1}}}
\providecommand{\Pubmed}[1]{\href{pmid:#1}{\path{#1}}}
\providecommand{\bibinfo}[2]{#2}
\ifx\xfnm\relax \def\xfnm[#1]{\unskip,\space#1}\fi
\bibitem[{Carlini and Wagner(2017)}]{carlini2017towards}
\bibinfo{author}{Carlini, N.}, \bibinfo{author}{Wagner, D.},
  \bibinfo{year}{2017}.
\newblock \bibinfo{title}{Towards evaluating the robustness of neural
  networks}, in: \bibinfo{booktitle}{2017 IEEE Symposium on Security and
  Privacy (SP)}, \bibinfo{organization}{IEEE}. pp. \bibinfo{pages}{39--57}.
\newblock \DOIprefix\doi{10.1109/SP.2017.49}.
\bibitem[{Das et~al.(2017)Das, Shanbhogue, Chen, Hohman, Chen, Kounavis and
  Chau}]{das2017keeping}
\bibinfo{author}{Das, N.}, \bibinfo{author}{Shanbhogue, M.},
  \bibinfo{author}{Chen, S.T.}, \bibinfo{author}{Hohman, F.},
  \bibinfo{author}{Chen, L.}, \bibinfo{author}{Kounavis, M.E.},
  \bibinfo{author}{Chau, D.H.}, \bibinfo{year}{2017}.
\newblock \bibinfo{title}{Keeping the bad guys out: Protecting and vaccinating
  deep learning with jpeg compression}.
\newblock \bibinfo{journal}{arXiv preprint arXiv:1705.02900} .
\bibitem[{Ginesu et~al.(2012)Ginesu, Pintus and Giusto}]{ginesu2012objective}
\bibinfo{author}{Ginesu, G.}, \bibinfo{author}{Pintus, M.},
  \bibinfo{author}{Giusto, D.D.}, \bibinfo{year}{2012}.
\newblock \bibinfo{title}{Objective assessment of the webp image coding
  algorithm}.
\newblock \bibinfo{journal}{Signal Processing: Image Communication}
  \bibinfo{volume}{27}, \bibinfo{pages}{867--874}.
\newblock \DOIprefix\doi{10.1016/j.image.2012.01.011}.
\bibitem[{Goodfellow et~al.(2014)Goodfellow, Shlens and
  Szegedy}]{Goodfellow2014Explaining}
\bibinfo{author}{Goodfellow, I.J.}, \bibinfo{author}{Shlens, J.},
  \bibinfo{author}{Szegedy, C.}, \bibinfo{year}{2014}.
\newblock \bibinfo{title}{Explaining and harnessing adversarial examples}.
\newblock \bibinfo{journal}{Computer Science} .
\bibitem[{Irons et~al.(2017)Irons, Gradden, Zhang, He, Barnes, Scott and
  McKone}]{irons2017face}
\bibinfo{author}{Irons, J.L.}, \bibinfo{author}{Gradden, T.},
  \bibinfo{author}{Zhang, A.}, \bibinfo{author}{He, X.},
  \bibinfo{author}{Barnes, N.}, \bibinfo{author}{Scott, A.F.},
  \bibinfo{author}{McKone, E.}, \bibinfo{year}{2017}.
\newblock \bibinfo{title}{Face identity recognition in simulated prosthetic
  vision is poorer than previously reported and can be improved by
  caricaturing}.
\newblock \bibinfo{journal}{Vision research} \bibinfo{volume}{137},
  \bibinfo{pages}{61--79}.
\newblock \DOIprefix\doi{10.1016/j.visres.2017.06.002}.
\bibitem[{Jia et~al.(2019)Jia, Wei, Cao and Foroosh}]{jia2019comdefend}
\bibinfo{author}{Jia, X.}, \bibinfo{author}{Wei, X.}, \bibinfo{author}{Cao,
  X.}, \bibinfo{author}{Foroosh, H.}, \bibinfo{year}{2019}.
\newblock \bibinfo{title}{Comdefend: An efficient image compression model to
  defend adversarial examples}, in: \bibinfo{booktitle}{Proceedings of the IEEE
  Conference on Computer Vision and Pattern Recognition}, pp.
  \bibinfo{pages}{6084--6092}.
\bibitem[{Krizhevsky et~al.(2012)Krizhevsky, Sutskever and
  Hinton}]{krizhevsky2012imagenet}
\bibinfo{author}{Krizhevsky, A.}, \bibinfo{author}{Sutskever, I.},
  \bibinfo{author}{Hinton, G.E.}, \bibinfo{year}{2012}.
\newblock \bibinfo{title}{Imagenet classification with deep convolutional
  neural networks}, in: \bibinfo{booktitle}{Advances in neural information
  processing systems}, pp. \bibinfo{pages}{1097--1105}.
\newblock \DOIprefix\doi{10.1145/3065386}.
\bibitem[{Kurakin et~al.(2016)Kurakin, Goodfellow and
  Bengio}]{kurakin2016adversarial}
\bibinfo{author}{Kurakin, A.}, \bibinfo{author}{Goodfellow, I.},
  \bibinfo{author}{Bengio, S.}, \bibinfo{year}{2016}.
\newblock \bibinfo{title}{Adversarial examples in the physical world}.
\newblock \bibinfo{journal}{arXiv preprint arXiv:1607.02533} .
\bibitem[{Lian and Shilei(2012)}]{lian2012webp}
\bibinfo{author}{Lian, L.}, \bibinfo{author}{Shilei, W.}, \bibinfo{year}{2012}.
\newblock \bibinfo{title}{Webp: A new image compression format based on vp8
  encoding}.
\newblock \bibinfo{journal}{Microcontrollers \& Embedded Systems}
  \bibinfo{volume}{3}.
\bibitem[{Liao et~al.(2018)Liao, Liang, Dong, Pang, Hu and
  Zhu}]{liao2018defense}
\bibinfo{author}{Liao, F.}, \bibinfo{author}{Liang, M.}, \bibinfo{author}{Dong,
  Y.}, \bibinfo{author}{Pang, T.}, \bibinfo{author}{Hu, X.},
  \bibinfo{author}{Zhu, J.}, \bibinfo{year}{2018}.
\newblock \bibinfo{title}{Defense against adversarial attacks using high-level
  representation guided denoiser}, in: \bibinfo{booktitle}{Proceedings of the
  IEEE Conference on Computer Vision and Pattern Recognition}, pp.
  \bibinfo{pages}{1778--1787}.
\bibitem[{Moosavi-Dezfooli et~al.(2016)Moosavi-Dezfooli, Fawzi and
  Frossard}]{moosavi2016deepfool}
\bibinfo{author}{Moosavi-Dezfooli, S.M.}, \bibinfo{author}{Fawzi, A.},
  \bibinfo{author}{Frossard, P.}, \bibinfo{year}{2016}.
\newblock \bibinfo{title}{Deepfool: a simple and accurate method to fool deep
  neural networks}, in: \bibinfo{booktitle}{Proceedings of the IEEE conference
  on computer vision and pattern recognition}, pp. \bibinfo{pages}{2574--2582}.
\newblock \DOIprefix\doi{10.1109/CVPR.2016.282}.
\bibitem[{Navarro et~al.(2018)Navarro, Deruyver and
  Parrend}]{navarro2018systematic}
\bibinfo{author}{Navarro, J.}, \bibinfo{author}{Deruyver, A.},
  \bibinfo{author}{Parrend, P.}, \bibinfo{year}{2018}.
\newblock \bibinfo{title}{A systematic survey on multi-step attack detection}.
\newblock \bibinfo{journal}{Computers \& Security} \bibinfo{volume}{76},
  \bibinfo{pages}{214--249}.
\newblock \DOIprefix\doi{10.1016/j.cose.2018.03.001}.
\bibitem[{Nguyen et~al.(2018)Nguyen, Le~Nguyen, Nguyen and
  Quan}]{nguyen2018auto}
\bibinfo{author}{Nguyen, M.H.}, \bibinfo{author}{Le~Nguyen, D.},
  \bibinfo{author}{Nguyen, X.M.}, \bibinfo{author}{Quan, T.T.},
  \bibinfo{year}{2018}.
\newblock \bibinfo{title}{Auto-detection of sophisticated malware using
  lazy-binding control flow graph and deep learning}.
\newblock \bibinfo{journal}{Computers \& Security} \bibinfo{volume}{76},
  \bibinfo{pages}{128--155}.
\newblock \DOIprefix\doi{10.1016/j.cose.2018.02.006}.
\bibitem[{Ni et~al.(2018)Ni, Qian and Zhang}]{ni2018malware}
\bibinfo{author}{Ni, S.}, \bibinfo{author}{Qian, Q.}, \bibinfo{author}{Zhang,
  R.}, \bibinfo{year}{2018}.
\newblock \bibinfo{title}{Malware identification using visualization images and
  deep learning}.
\newblock \bibinfo{journal}{Computers \& Security} \bibinfo{volume}{77},
  \bibinfo{pages}{871--885}.
\newblock \DOIprefix\doi{10.1016/j.cose.2018.04.005}.
\bibitem[{Ozah and Kolokolova(2019)}]{ozah2019compression}
\bibinfo{author}{Ozah, N.}, \bibinfo{author}{Kolokolova, A.},
  \bibinfo{year}{2019}.
\newblock \bibinfo{title}{Compression improves image classification accuracy},
  in: \bibinfo{booktitle}{Canadian Conference on Artificial Intelligence},
  \bibinfo{organization}{Springer}. pp. \bibinfo{pages}{525--530}.
\bibitem[{Papernot et~al.(2017)Papernot, McDaniel, Goodfellow, Jha, Celik and
  Swami}]{papernot2017practical}
\bibinfo{author}{Papernot, N.}, \bibinfo{author}{McDaniel, P.},
  \bibinfo{author}{Goodfellow, I.}, \bibinfo{author}{Jha, S.},
  \bibinfo{author}{Celik, Z.B.}, \bibinfo{author}{Swami, A.},
  \bibinfo{year}{2017}.
\newblock \bibinfo{title}{Practical black-box attacks against machine
  learning}, in: \bibinfo{booktitle}{Proceedings of the 2017 ACM on Asia
  conference on computer and communications security},
  \bibinfo{organization}{ACM}. pp. \bibinfo{pages}{506--519}.
\newblock \DOIprefix\doi{10.1145/3052973.3053009}.
\bibitem[{Pedregosa et~al.(2011)Pedregosa, Varoquaux, Gramfort, Michel,
  Thirion, Grisel, Blondel, Prettenhofer, Weiss, Dubourg
  et~al.}]{pedregosa2011scikit}
\bibinfo{author}{Pedregosa, F.}, \bibinfo{author}{Varoquaux, G.},
  \bibinfo{author}{Gramfort, A.}, \bibinfo{author}{Michel, V.},
  \bibinfo{author}{Thirion, B.}, \bibinfo{author}{Grisel, O.},
  \bibinfo{author}{Blondel, M.}, \bibinfo{author}{Prettenhofer, P.},
  \bibinfo{author}{Weiss, R.}, \bibinfo{author}{Dubourg, V.}, et~al.,
  \bibinfo{year}{2011}.
\newblock \bibinfo{title}{Scikit-learn: Machine learning in python}.
\newblock \bibinfo{journal}{Journal of machine learning research}
  \bibinfo{volume}{12}, \bibinfo{pages}{2825--2830}.
\bibitem[{Raid et~al.(2014)Raid, Khedr, El-Dosuky and Ahmed}]{Raid2014Jpeg}
\bibinfo{author}{Raid, A.M.}, \bibinfo{author}{Khedr, W.M.},
  \bibinfo{author}{El-Dosuky, M.A.}, \bibinfo{author}{Ahmed, W.},
  \bibinfo{year}{2014}.
\newblock \bibinfo{title}{Jpeg image compression using discrete cosine
  transform - a survey}.
\newblock \bibinfo{journal}{International Journal of Computer Science \&
  Engineering Survey} \bibinfo{volume}{5}, \bibinfo{pages}{39--47}.
\bibitem[{Russell and Norvig(2016)}]{russell2016artificial}
\bibinfo{author}{Russell, S.J.}, \bibinfo{author}{Norvig, P.},
  \bibinfo{year}{2016}.
\newblock \bibinfo{title}{Artificial intelligence: a modern approach}.
\newblock \bibinfo{publisher}{Malaysia; Pearson Education Limited}.
\bibitem[{Singh et~al.(2007)Singh, Kumar and Verma}]{singh2007reduction}
\bibinfo{author}{Singh, S.}, \bibinfo{author}{Kumar, V.},
  \bibinfo{author}{Verma, H.}, \bibinfo{year}{2007}.
\newblock \bibinfo{title}{Reduction of blocking artifacts in jpeg compressed
  images}.
\newblock \bibinfo{journal}{Digital signal processing} \bibinfo{volume}{17},
  \bibinfo{pages}{225--243}.
\newblock \DOIprefix\doi{10.1016/j.dsp.2005.08.003}.
\bibitem[{Song et~al.(2017)Song, Kim, Nowozin, Ermon and
  Kushman}]{song2017pixeldefend}
\bibinfo{author}{Song, Y.}, \bibinfo{author}{Kim, T.},
  \bibinfo{author}{Nowozin, S.}, \bibinfo{author}{Ermon, S.},
  \bibinfo{author}{Kushman, N.}, \bibinfo{year}{2017}.
\newblock \bibinfo{title}{Pixeldefend: Leveraging generative models to
  understand and defend against adversarial examples}.
\newblock \bibinfo{journal}{arXiv preprint arXiv:1710.10766} .
\bibitem[{Szegedy et~al.(2013)Szegedy, Zaremba, Sutskever, Bruna, Erhan,
  Goodfellow and Fergus}]{Szegedy2013Intriguing}
\bibinfo{author}{Szegedy, C.}, \bibinfo{author}{Zaremba, W.},
  \bibinfo{author}{Sutskever, I.}, \bibinfo{author}{Bruna, J.},
  \bibinfo{author}{Erhan, D.}, \bibinfo{author}{Goodfellow, I.},
  \bibinfo{author}{Fergus, R.}, \bibinfo{year}{2013}.
\newblock \bibinfo{title}{Intriguing properties of neural networks}.
\newblock \bibinfo{journal}{Computer Science} .
\bibitem[{Thang and Matsui(2019)}]{thang2019image}
\bibinfo{author}{Thang, D.D.}, \bibinfo{author}{Matsui, T.},
  \bibinfo{year}{2019}.
\newblock \bibinfo{title}{Image transformation can make neural networks more
  robust against adversarial examples}.
\newblock \bibinfo{journal}{arXiv preprint arXiv:1901.03037} .
\bibitem[{Tram{\`e}r et~al.(2017)Tram{\`e}r, Kurakin, Papernot, Goodfellow,
  Boneh and McDaniel}]{tramer2017ensemble}
\bibinfo{author}{Tram{\`e}r, F.}, \bibinfo{author}{Kurakin, A.},
  \bibinfo{author}{Papernot, N.}, \bibinfo{author}{Goodfellow, I.},
  \bibinfo{author}{Boneh, D.}, \bibinfo{author}{McDaniel, P.},
  \bibinfo{year}{2017}.
\newblock \bibinfo{title}{Ensemble adversarial training: Attacks and defenses}.
\newblock \bibinfo{journal}{arXiv preprint arXiv:1705.07204} .
\bibitem[{Wang et~al.(2017)Wang, Wang and Zhou}]{wang2017bimodal}
\bibinfo{author}{Wang, J.}, \bibinfo{author}{Wang, G.}, \bibinfo{author}{Zhou,
  M.}, \bibinfo{year}{2017}.
\newblock \bibinfo{title}{Bimodal vein data mining via cross-selected-domain
  knowledge transfer}.
\newblock \bibinfo{journal}{IEEE Transactions on Information Forensics and
  Security} \bibinfo{volume}{13}, \bibinfo{pages}{733--744}.
\bibitem[{Xie et~al.(2017)Xie, Wang, Zhang, Ren and Yuille}]{xie2017mitigating}
\bibinfo{author}{Xie, C.}, \bibinfo{author}{Wang, J.}, \bibinfo{author}{Zhang,
  Z.}, \bibinfo{author}{Ren, Z.}, \bibinfo{author}{Yuille, A.},
  \bibinfo{year}{2017}.
\newblock \bibinfo{title}{Mitigating adversarial effects through
  randomization}.
\newblock \bibinfo{journal}{arXiv preprint arXiv:1711.01991} .
\bibitem[{Xing et~al.(2016)Xing, Zhang and Shang}]{Xing2016Deep}
\bibinfo{author}{Xing, H.}, \bibinfo{author}{Zhang, G.},
  \bibinfo{author}{Shang, M.}, \bibinfo{year}{2016}.
\newblock \bibinfo{title}{Deep learning}.
\newblock \bibinfo{journal}{International Journal of Semantic Computing}
  \bibinfo{volume}{10}, \bibinfo{pages}{417--439}.
\newblock \DOIprefix\doi{10.1142/S1793351X16500045}.
\bibitem[{Zhong et~al.(2017)Zhong, Lei, Cao, Fan and Li}]{zhong2017class}
\bibinfo{author}{Zhong, Z.}, \bibinfo{author}{Lei, M.}, \bibinfo{author}{Cao,
  D.}, \bibinfo{author}{Fan, J.}, \bibinfo{author}{Li, S.},
  \bibinfo{year}{2017}.
\newblock \bibinfo{title}{Class-specific object proposals re-ranking for object
  detection in automatic driving}.
\newblock \bibinfo{journal}{Neurocomputing} \bibinfo{volume}{242},
  \bibinfo{pages}{187--194}.
\newblock \DOIprefix\doi{10.1016/j.neucom.2017.02.068}.

\end{thebibliography}
\end{document}